\documentclass[sigconf]{acmart}
\usepackage{amsmath}
\usepackage{booktabs}
\usepackage{graphicx}
\usepackage{xcolor}
\usepackage{url}

\definecolor{improvementgreen}{RGB}{0,120,70}
\definecolor{deteriorationred}{RGB}{190,35,45}
\newcommand{\goodstars}[1]{\textcolor{improvementgreen}{\textbf{#1}}}
\newcommand{\badstars}[1]{\textcolor{deteriorationred}{\textbf{#1}}}

\renewcommand\footnotetextcopyrightpermission[1]{}
\acmConference[Preprint]{}{Under Review}{-}

\acmYear{2026}
\copyrightyear{2026}

\begin{document}

\title{Crossing-Free Probabilistic K-Line Forecasts Without Retraining}

\author{Runyao Yu}
\authornote{Runyao Yu is also affiliated with Delft University of Technology and the AIT Austrian Institute of Technology.}
\affiliation{%
  \institution{London Business School}
  \city{London}
  \country{United Kingdom}
}
\email{ryu@london.edu}

\author{Yuchen Tao}
\affiliation{%
  \institution{RWTH Aachen University}
  \city{Aachen}
  \country{Germany}
}
\email{yuchen.tao@rwth-aachen.de}

\author{Yujie Chen}
\affiliation{%
  \institution{The Chinese University of Hong Kong}
  \city{Shenzhen}
  \country{China}
}
\email{chenyujie@cuhk.edu.cn}

\author{Wentao Wang}
\affiliation{%
  \institution{University of Technology Sydney}
  \city{Sydney}
  \country{Australia}
}
\email{wentao.wang@student.uts.edu.au}

\author{Derek W. Bunn}
\affiliation{%
  \institution{London Business School}
  \city{London}
  \country{United Kingdom}
}
\email{dbunn@london.edu}

\renewcommand{\shortauthors}{Yu et al.}

\begin{abstract}
Probabilistic K-line forecasting describes uncertainty in four complementary prices, namely open--high--low--close (OHLC). 
However, it introduces two consistency problems: quantile crossing and K-line crossing. Quantile crossing occurs when a higher-quantile forecast falls below a lower-quantile forecast, while K-line crossing occurs when the forecast low exceeds the open or close, or the forecast high falls below the open or close.
Existing solutions generally address only one problem through output reordering, specialized architectures, or penalized training objectives. We propose K-line--Quantile Sequential Projection (KQSP), a parameter-free and training-free reconciliation method applicable to forecasts produced by any model. 
Compared with other crossing solutions, KQSP preserves predictive accuracy while producing substantially smaller corrections to the original forecasts.
To mitigate model bias, we evaluate KQSP using various models, including pretrained foundation models. KQSP reduces both quantile and K-line crossing rates to zero for all test data undertaken. 
These results show that probabilistic K-line consistency can be enforced independently of forecast generation and without retraining. 
\end{abstract}



\maketitle



\section{Introduction}
\label{sec:introduction}

Probabilistic forecasts are particularly valuable in uncertain financial markets as they describe possible outcomes and tail risks
rather than providing only one expected value~\cite{gneiting2014probabilistic}. However, most probabilistic financial forecasting studies focus on a single price target~\cite{barunik2026forecasting,taylor1999quantile,bunn2016analysis}, such as index level or close price~\cite{jang2024taft,kim2025iknet}.
K-line forecasting, also known as candlestick or open--high--low--close (OHLC) forecasting, predicts four complementary price landmarks~\cite{huang2024structural,huang2023transformer}. The open price reflects the valuation at the beginning of a trading period, the high and low prices determine the realized trading range and extreme movements, and the close price summarizes the terminal valuation. Joint OHLC forecasts can consequently support entry and exit decisions, stop-loss placement, intraperiod risk assessment, and trading strategies that cannot be constructed from a single price index alone.

However, existing K-line forecasting research remains predominantly pointwise. Structural VAR and VECM models forecast one OHLC vector for each period~\cite{huang2024structural}, while Transformer-based methods similarly produce point forecasts for the four prices~\cite{huang2023transformer}. 
Conversely, most probabilistic financial forecasting studies estimate the distribution of only one target~\cite{yang2024cqvae,barunik2026forecasting}.
Probabilistic K-line forecasting combines these two settings by predicting multiple quantiles for all four prices, thereby describing the uncertainty of the opening price, closing price, and trading range simultaneously. 
Although financial econometric models commonly forecast returns as asset price levels are often nonstationary, we retain price levels to align with the existing K-line forecasting literature and preserve the direct interpretation of OHLC forecasts.
Kronos represents a rare recent attempt to generate K-line paths~\cite{shi2026kronos}. Nevertheless, to our knowledge, the simultaneous reconciliation of probabilistic K-line forecasts has not been studied as a model-agnostic and training-free problem.

\begin{figure*}[t]
  \centering
  \includegraphics[width=1\textwidth]{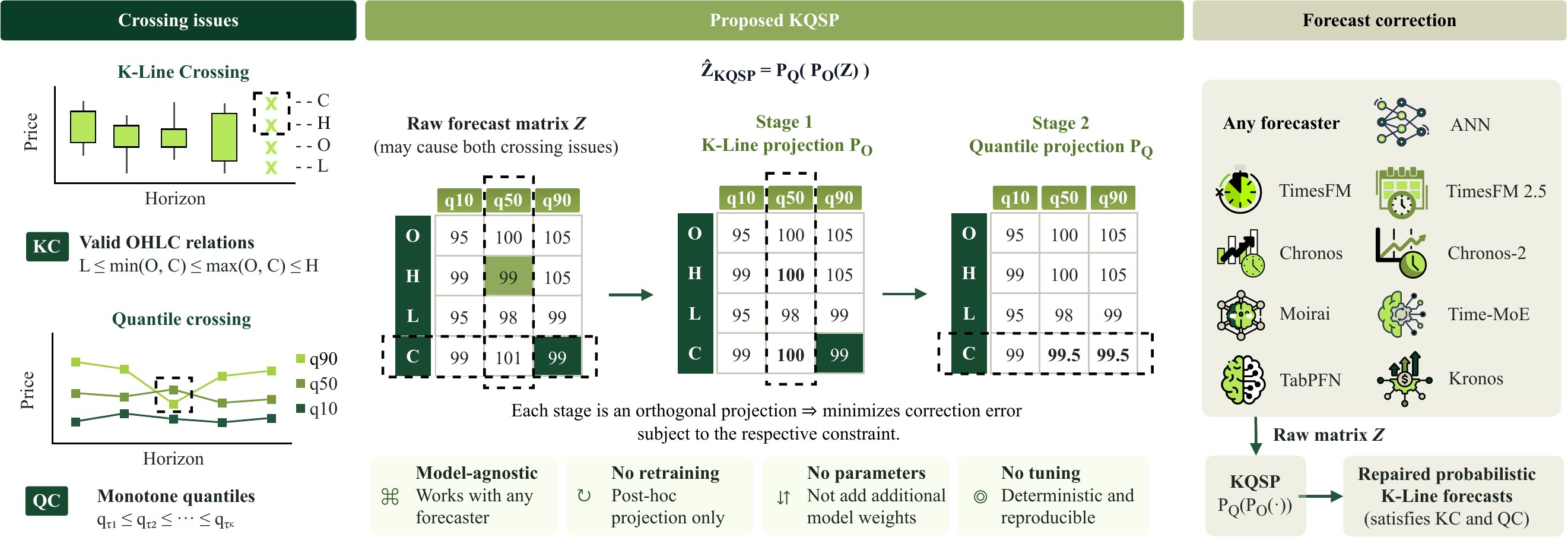}
  \caption{Overview of the work. Left: illustration of the two crossing issues. Middle: a numerical example of KQSP, which sequentially applies the K-line projection and quantile projection to repair the raw forecast. Right: model-agnostic application of KQSP to probabilistic K-line forecasts generated by various models.}
  \label{fig:overview}
  \Description{Diagram showing the two crossing problems, the two-stage KQSP correction process, and its application to different forecasting models.}
\end{figure*}

Probabilistic K-line forecasts introduce two distinct consistency problems. As illustrated in Fig.~\ref{fig:overview}, quantile crossing occurs when a higher quantile falls below a lower quantile; K-line crossing occurs when the predicted high is lower than the open or close, or the predicted low is higher than the open or close. These violations can occur simultaneously, as a probabilistic K-line forecast must preserve both the quantile order of every feature and the OHLC relations at every quantile level.

\begin{figure*}[t]
  \centering
  \includegraphics[width=1\textwidth]{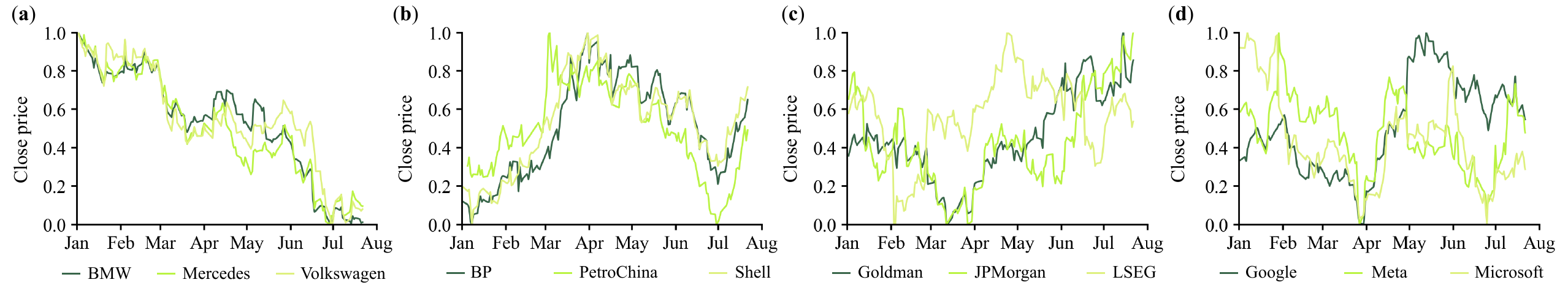}
  \caption{Min--max-scaled close prices over the common period from January 1 to July 23, 2026. The panels show three stocks from each category: \textbf{(a)} Automobile, \textbf{(b)} Energy, \textbf{(c)} Finance, and \textbf{(d)} Technology.}
  \label{fig:data-categories}
  \Description{Four line charts comparing scaled closing prices for stocks from the automobile, energy, finance, and technology categories.}
\end{figure*}

K-line crossing has received comparatively limited attention. Existing solutions incorporate OHLC relations through model-specific designs and penalty terms in the training loss~\cite{huang2023transformer}. Quantile crossing has been studied more extensively through post-hoc reordering~\cite{chernozhukov2010quantile}, penalized quantile-regression objectives~\cite{shen2024noncrossing}, and hierarchical multi-quantile heads~\cite{yu2026pricefmfoundationmodelprobabilistic, yu2026orderfusionencodingorderbookendtoend, YU2026105083}. However, these methods are designed for one constraint family and do not independently guarantee both quantile and K-line consistency. Moreover, training-time approaches require model modification, retraining, or penalty selection, whereas reordering can alter the original quantile assignments and change forecasts more than necessary.

In this work, we propose K-line--Quantile Sequential Projection (KQSP), a training-free method that reconciles any existing probabilistic K-line forecast. KQSP first applies the minimum-distance K-line projection at each quantile level and then applies the minimum-distance quantile projection to each OHLC feature, which will be detailed in Section~\ref{sec:method}. KQSP introduces no additional parameter, hyperparameter, loss penalty, or retraining requirement. We evaluate KQSP on various stocks using forecasts from a fully trained Artificial Neural Network (ANN) and eight zero-shot foundation models, and compare it with multiple crossing solutions. KQSP eliminates both crossing types for every evaluated case without compromising predictive performance. Instead, these predictive metrics are frequently improved with statistical significance, while KQSP changes the original forecasts less than other crossing solutions.
The main contributions are threefold:
\begin{itemize}
    \item We propose K-line--Quantile Sequential Projection (KQSP), a model-agnostic method that removes quantile and K-line crossings, without introducing parameters, hyperparameters, loss penalties, or retraining.

    \item We systematically compare KQSP with multiple crossing solutions. KQSP achieves comparable predictive accuracy to the other methods while producing significantly smaller corrections.

    \item We evaluate KQSP across a fully trained ANN and various zero-shot foundation models. KQSP eliminates all remaining crossings and consistently improves AQL, MAE, and RMSE, demonstrating robustness across forecasting models with substantially different architectures.
\end{itemize}

\section{Preliminary}
\label{sec:preliminary}

\subsection{Quantile Crossing}
Let $\mathcal T=\{\tau_1<\cdots<\tau_Q\}$ be $Q$ ordered quantile levels, let $\mathcal F=\{O,H,L,C\}$ denote open, high, low, and close, and let $\mathbf Z=[z_{qf}]\in\mathbb R^{Q\times4}$ be one raw probabilistic forecast, where $z_{qf}$ is the forecast at level $\tau_q$ for feature $f\in\mathcal F$.
Quantile consistency requires every feature-specific vector to belong to the monotone cone
\begin{equation}
  \mathcal C_{\mathrm Q}=\{\mathbf v\in\mathbb R^Q:v_1\leq\cdots\leq v_Q\}.
  \label{eq:quantile-constraint}
\end{equation}
A quantile crossing occurs when $z_{qf}>z_{q+1,f}$ for at least one adjacent pair. Independent unconstrained outputs can violate this order under finite data and imperfect optimization.
For example, at levels $(0.1,0.5,0.9)$, open forecasts $(95,105,\textbf{100})$ cross because the $0.9$-quantile forecast $100$ is lower than the median forecast $105$.

\begin{table*}[t]
\centering
\scriptsize
\setlength{\tabcolsep}{1.8pt}
\caption{\textbf{Data set summary.} Mean and standard deviation of daily OHLC prices in native units.}
\resizebox{\textwidth}{!}{%
\begin{tabular}{lllllrrrrrrrr}
\toprule
& & & & 
& \multicolumn{2}{c}{Open}
& \multicolumn{2}{c}{High}
& \multicolumn{2}{c}{Low}
& \multicolumn{2}{c}{Close} \\
\cmidrule(lr){6-7}
\cmidrule(lr){8-9}
\cmidrule(lr){10-11}
\cmidrule(lr){12-13}
Stock & Category & Exchange & Currency & Ticker
& Mean & Std.
& Mean & Std.
& Mean & Std.
& Mean & Std. \\
\midrule
BMW
& Automobile
& Xetra & EUR & BMW
& 81.05 & 12.56
& 81.90 & 12.57
& 80.16 & 12.53
& 81.04 & 12.57 \\

Mercedes
& Automobile
& Xetra & EUR & MBG
& 54.46 & 11.27
& 55.05 & 11.32
& 53.85 & 11.21
& 54.45 & 11.27 \\

Volkswagen
& Automobile
& Xetra & EUR & VOW
& 153.75 & 50.35
& 155.69 & 51.26
& 151.67 & 49.23
& 153.62 & 50.27 \\
\midrule

BP
& Energy
& LSE & GBX & BP.
& 441.95 & 85.97
& 446.89 & 85.88
& 437.12 & 85.80
& 441.86 & 85.90 \\

PetroChina
& Energy
& HKEX & HKD & 00857
& 7.14 & 1.87
& 7.22 & 1.91
& 7.07 & 1.85
& 7.15 & 1.88 \\

Shell
& Energy
& LSE & GBX & SHEL
& 2251.40 & 514.40
& 2274.34 & 515.77
& 2229.66 & 513.93
& 2251.21 & 515.20 \\
\midrule

Goldman
& Finance
& NYSE & USD & GS
& 367.03 & 206.59
& 371.23 & 209.41
& 362.96 & 203.84
& 367.20 & 206.76 \\

JPMorgan
& Finance
& NYSE & USD & JPM
& 153.79 & 68.84
& 155.29 & 69.55
& 152.34 & 68.19
& 153.85 & 68.90 \\

LSEG
& Finance
& LSE & GBX & LSEG
& 7161.82 & 2405.01
& 7238.58 & 2429.66
& 7083.68 & 2377.85
& 7161.80 & 2403.31 \\
\midrule

Google
& Technology
& Nasdaq & USD & GOOG
& 117.48 & 76.88
& 118.84 & 77.89
& 116.24 & 75.94
& 117.58 & 76.99 \\

Meta
& Technology
& Nasdaq & USD & META
& 308.56 & 185.62
& 312.49 & 187.71
& 304.56 & 183.09
& 308.59 & 185.39 \\

Microsoft
& Technology
& Nasdaq & USD & MSFT
& 249.07 & 137.73
& 251.42 & 138.87
& 246.57 & 136.44
& 249.10 & 137.69 \\
\bottomrule
\end{tabular}%
}
\label{datasummarytab}
\end{table*}

\subsection{K-Line Crossing}
For quantile index $q$, let $\mathbf z_q=(z_{qf})_{f\in\mathcal F}$ denote its OHLC forecast.
K-line consistency requires this vector to belong to
\begin{equation}
  \begin{split}
  \mathcal C_{\mathrm O}=\{(O,H,L,C)\in\mathbb R^4:\;&L\leq\min(O,C),\\
  &\max(O,C)\leq H\}.
  \end{split}
  \label{eq:ohlc-constraint}
\end{equation}
A K-line crossing occurs when $z_{qH}<\max\{z_{qO},z_{qC}\}$ or $z_{qL}>\min\{z_{qO},z_{qC}\}$. Independent feature outputs need not preserve these semantic bounds.
For example, $(100,\textbf{99},98,101)$ in $(O,H,L,C)$ order crosses because high $99$ is below close $101$.

\section{Data}
\label{sec:data}

We use 30,174 daily OHLCV observations, where V denotes trading volume, from 12 stocks across four categories.
The data span July 23, 2016 to July 23, 2026, with 2,427--2,537 observations per stock. 
Each eligible sample uses a historical OHLCV window to predict the next-day OHLC vector.
Samples are ordered chronologically and split without shuffling into 70\% training, 15\% validation, and 15\% testing subsets.
All input and output scalers are fitted on training data only.

Table~\ref{datasummarytab} reports descriptive statistics of the OHLC prices in their native units. For visual comparability and to reveal the overall temporal trends, each category panel in Fig.~\ref{fig:data-categories} rescales every stock's 2026 close prices to $[0,1]$ over the displayed period. Together, the summary table and figure demonstrate substantial diversity in price scale, temporal shape, and business domain, providing heterogeneous cases for evaluating forecast reconciliation.

\section{Method}
\label{sec:method}

\subsection{Backbone-Agnostic Forecast Interface}
The probabilistic K-line forecasts $\mathbf Z=[z_{qf}]\in\mathbb R^{Q\times4}$ may come from a fully trained ANN, a zero-shot foundation model (FM), or any other backbone with the same output interface.
For an ANN, a shared representation $\mathbf h=f_{\boldsymbol\theta}(\mathbf x)$ is mapped by its existing output layer to $\mathbf Z$, where $\mathbf x$ is the historical input and $\boldsymbol\theta$ denotes the backbone parameters.
Our primary goal is \textbf{not} to optimize the backbone, engineer features, or construct optimal factors for predictive accuracy.
Instead, we isolate whether quantile and K-line crossings can be removed from existing density forecasts, regardless of whether their native predictive performance is strong or weak.

\subsection{K-line--Quantile Sequential Projection (KQSP)}
\label{sec:oqsp}

KQSP is a parameter-free sequential projection method that makes probabilistic K-line forecasts $\mathbf Z=[z_{qf}]\in\mathbb R^{Q\times4}$ crossing-free. Each row of $\mathbf Z$ contains the OHLC forecasts at one quantile level, while each column contains all quantile forecasts for one OHLC feature. KQSP first corrects each row to satisfy the K-line constraints and then corrects each column to satisfy the quantile order, using the minimum-distance correction at each stage. The projection is
\begin{equation}
\widehat{\mathbf Z}^{\mathrm{KQSP}}
=
\mathcal P_{\mathrm Q}\!\left(
\mathcal P_{\mathrm O}(\mathbf Z)
\right),
\label{eq:oqsp}
\end{equation}
where $\mathcal P_{\mathrm O}$ denotes the OHLC projection, $\mathcal P_{\mathrm Q}$ denotes the quantile projection, and $\widehat{\mathbf Z}^{\mathrm{KQSP}}$ is the final forecast.

The size of a correction is measured by
\begin{equation}
E_{\mathrm{corr}}(\widehat{\mathbf Z},\mathbf Z)
=
\sum_{q=1}^{Q}
\sum_{f\in\mathcal F}
(\widehat z_{qf}-z_{qf})^2.
\label{eq:correction-error}
\end{equation}
A smaller value means that the corrected forecast remains closer to the original forecast, which is critical when the original model is considered reliable and only the minimum correction required to remove crossings is desired.

\textbf{Stage 1: K-line projection.}
For each quantile level $q$, KQSP finds the valid OHLC vector closest to the original row:
\begin{equation}
\mathbf c_q^{\star}
=
\arg\min_{\mathbf c\in\mathcal C_{\mathrm O}}
\|\mathbf c-\mathbf z_q\|_2^2,
\label{eq:ohlc-projection}
\end{equation}
where $\mathcal C_{\mathrm O}$ contains all vectors satisfying $H\geq O$, $H\geq C$, $L\leq O$, and $L\leq C$. Because there are only four relations, KQSP checks every possible combination in which one or more relations become equalities and selects the valid result with the smallest correction.

For example, consider $(O,H,L,C)=(100,\textbf{99},98,101)$. The high forecast is below both the open and close forecasts. One possible correction is to raise $H$ from $99$ to $101$, producing $(100,\textbf{101},98,101)$ with squared correction $(101-99)^2=4$. KQSP instead makes a smaller joint correction: it raises $H$ from $99$ to $100$ and lowers $C$ from $101$ to $100$. It therefore returns $\mathbf c_q^{\star}=(100,\textbf{100},98,\textbf{100})$ with squared correction $(100-99)^2+(100-101)^2=2$. Thus, KQSP satisfies all OHLC relations while changing the original forecast less than the one-sided correction.

\textbf{Stage 2: quantile projection.}
After correcting all rows, let $\mathbf c_{:f}^{\star}$ contain the forecasts of feature $f$ across all quantile levels. KQSP finds the closest nondecreasing sequence:
\begin{equation}
\widehat{\mathbf z}_{:f}^{\mathrm{KQSP}}
=
\arg\min_{\mathbf v\in\mathcal C_{\mathrm Q}}
\|\mathbf v-\mathbf c_{:f}^{\star}\|_2^2,
\label{eq:quantile-projection}
\end{equation}
where $\mathcal C_{\mathrm Q}$ contains all sequences satisfying
$v_1\leq\cdots\leq v_Q$. KQSP checks the quantiles from low to high. Whenever consecutive values decrease, it replaces the violating group with its mean and continues until the complete sequence is nondecreasing.

For example, consider quantile levels $(\tau_1,\tau_2,\tau_3)=(0.1,0.5,0.9)$ and the crossed sequence $\mathbf c_{:f}^{\star}=(99,100,\textbf{99})$. The forecasts at $\tau_2$ and $\tau_3$ are incorrectly ordered because $99<100$. One possible correction is to raise the $\tau_3$ forecast from $99$ to $100$, producing $(99,100,\textbf{100})$ with squared correction $(99-100)^2=1$. KQSP instead makes a smaller joint correction: it lowers the $\tau_2$ forecast from $100$ to $99.5$ and raises the $\tau_3$ forecast from $99$ to $99.5$. It therefore returns $\widehat{\mathbf z}_{:f}^{\mathrm{KQSP}}=(99,\textbf{99.5},\textbf{99.5})$ with squared correction $(100-99.5)^2+(99-99.5)^2=0.5$. Thus, KQSP restores the quantile order while changing the original forecasts less than the one-sided correction.
The exemplary correction errors are used in Section~\ref{sec:baselines} to explain the intuition behind KQSP and compare it with alternative crossing solutions.

\textbf{Preservation of K-line consistency.}
Stage~2 does not reintroduce the K-line crossings removed in Stage~1. After Stage~1, the forecasts satisfy
$c_{qL}^{\star}\leq c_{qO}^{\star}$,
$c_{qL}^{\star}\leq c_{qC}^{\star}$,
$c_{qO}^{\star}\leq c_{qH}^{\star}$, and
$c_{qC}^{\star}\leq c_{qH}^{\star}$
at every quantile level $q$. Stage~2 applies the same minimum-distance projection to each feature column. This projection preserves the order between columns: if one column is no greater than another at every quantile level before projection, it remains no greater afterward. Therefore,
\begin{equation}
\widehat z_{qL}^{\mathrm{KQSP}}
\leq
\min\!\left\{
\widehat z_{qO}^{\mathrm{KQSP}},
\widehat z_{qC}^{\mathrm{KQSP}}
\right\},
\qquad
\widehat z_{qH}^{\mathrm{KQSP}}
\geq
\max\!\left\{
\widehat z_{qO}^{\mathrm{KQSP}},
\widehat z_{qC}^{\mathrm{KQSP}}
\right\},
\quad \forall q.
\label{eq:ohlc-order-preserved}
\end{equation}
Hence, Stage~2 enforces quantile order while preserving the K-line consistency obtained in Stage~1.

\textbf{Method Characteristics.}
KQSP introduces no additional parameter, hyperparameter, penalty term, or loss modification and requires no retraining. It can be applied directly to probabilistic K-line forecasts produced by any forecasting model, regardless of its architecture or training procedure. Because KQSP operates only on the forecast matrix, the original model and its predictive process remain unchanged.

\section{Baselines}
\label{sec:baselines}

\subsection{Crossing Solutions}

\noindent\textbf{No Constraint.}
This baseline returns the raw forecast without correction:
\begin{equation}
\widehat{\mathbf Z}^{\mathrm{NC}}
=
\mathbf Z.
\label{eq:no-constraint}
\end{equation}

\noindent\textbf{Cumulative Maximum.}
This baseline first corrects each OHLC row while keeping its open and close forecasts unchanged:
\begin{equation}
\overline{\mathbf z}_{q}^{\mathrm{CM}}
=
\left(
z_{qO},
\max\{z_{qH},z_{qO},z_{qC}\},
\min\{z_{qL},z_{qO},z_{qC}\},
z_{qC}
\right).
\label{eq:cummax-ohlc}
\end{equation}
It then corrects each feature column by replacing every forecast with the largest value observed at that or any lower quantile level:
\begin{equation}
\widehat z_{qf}^{\mathrm{CM}}
=
\max_{1\leq j\leq q}
\overline z_{jf}^{\mathrm{CM}},
\label{eq:cummax-quantile}
\end{equation}
where $j$ indexes the quantile levels up to $q$.


For the OHLC example $(100,\textbf{99},98,101)$ and the quantile example $(99,100,\textbf{99})$ introduced in Section~\ref{sec:oqsp}, Table~\ref{tab:correction-examples} summarizes the corrected forecasts and corresponding correction errors. The cumulative maximum method increases the correction error by $100\%$ compared with KQSP.

\begin{table}[t]
\centering
\footnotesize
\caption{Corrections for the K-line example $(100,99,98,101)$ and the quantile example $(99,100,99)$. Bold values indicate corrected forecasts.}
\label{tab:correction-examples}
\renewcommand{\arraystretch}{1.15}
\begin{tabular}{llll}
\toprule
Method
& K-line forecast
& Quantile forecast
& Correction error \\
\midrule
KQSP
& $(100,\textbf{100},98,\textbf{100})$
& $(99,\textbf{99.5},\textbf{99.5})$
& $2+0.5=2.5$ \\

Cumulative Max.
& $(100,\textbf{101},98,101)$
& $(99,100,\textbf{100})$
& $4+1=5$ \\

Reordering
& $(\textbf{99},\textbf{101},98,\textbf{100})$
& $(99,\textbf{99},\textbf{100})$
& $6+2=8$ \\

Joint Projection
& $(100,\textbf{100},98,\textbf{100})$
& $(99,\textbf{99.5},\textbf{99.5})$
& $2+0.5=2.5$ \\
\bottomrule
\end{tabular}
\end{table}

\noindent\textbf{Reordering.}
This baseline corrects an OHLC row by sorting its four values. Let
$a_{q(1)}\leq a_{q(2)}\leq a_{q(3)}\leq a_{q(4)}$
denote the sorted values of $\mathbf z_q$. The minimum is assigned to low, the maximum to high, and the two middle values to open and close while retaining their original direction:
\begin{equation}
\overline{\mathbf z}_{q}^{\mathrm{RE}}
=
\begin{cases}
\left(a_{q(2)},a_{q(4)},a_{q(1)},a_{q(3)}\right),
& z_{qO}\leq z_{qC},\\
\left(a_{q(3)},a_{q(4)},a_{q(1)},a_{q(2)}\right),
& z_{qO}>z_{qC}.
\end{cases}
\label{eq:reordering-ohlc}
\end{equation}
It then sorts each feature column in ascending quantile order:
\begin{equation}
\widehat{\mathbf z}_{:f}^{\mathrm{RE}}
=
\operatorname{sort}_{\uparrow}
\left(
\overline z_{1f}^{\mathrm{RE}},
\ldots,
\overline z_{Qf}^{\mathrm{RE}}
\right).
\label{eq:reordering-quantile}
\end{equation}


Although Reordering removes both crossing types, it permutes the predicted values instead of minimizing their displacement. Consequently, a value originally produced for one OHLC feature or quantile level may be reassigned to another, breaking the association between the input and output. As shown in Table~\ref{tab:correction-examples}, Reordering yields the highest total correction error for these examples.

\noindent\textbf{Hierarchical.}
The hierarchical head was proposed in \cite{yu2026orderfusionencodingorderbookendtoend}.
Let $\mathbf U=[u_{qf}]\in\mathbb R^{Q\times4}$ denote the native output of the hierarchical model, where $q$ indexes the quantile level and $f\in\mathcal F$ indexes the OHLC feature. The non-negative residual is
\begin{equation}
r_{qf}=|u_{qf}|.
\label{eq:hierarchical-residual}
\end{equation}

Let $m$ denote the median-quantile index satisfying $\tau_m=0.5$. Hierarchical$^1$ uses the median output as its starting point and recursively adds or subtracts the residuals:
\begin{equation}
\widehat z_{qf}^{\mathrm{H1}}
=
\begin{cases}
u_{mf}, & q=m,\\
\widehat z_{q-1,f}^{\mathrm{H1}}+r_{qf}, & q>m,\\
\widehat z_{q+1,f}^{\mathrm{H1}}-r_{qf}, & q<m.
\end{cases}
\label{eq:hierarchical-quantile}
\end{equation}
This construction enforces quantile order but does not enforce the OHLC relations.

Inspired by Hierarchical$^1$, 
Hierarchical$^2$ interprets $u_{qO}$ and $u_{qC}$ as the open and close forecasts and uses the remaining outputs as non-negative high and low residuals:
\begin{equation}
\widehat{\mathbf z}_{q}^{\mathrm{H2}}
=
\left(
u_{qO},
\max\{u_{qO},u_{qC}\}+r_{qH},
\min\{u_{qO},u_{qC}\}-r_{qL},
u_{qC}
\right).
\label{eq:hierarchical-ohlc}
\end{equation}
This construction enforces the OHLC relations but does not enforce quantile order. 

Hierarchical$^3$ combines the recursive quantile construction of Hierarchical$^1$ with the OHLC construction of Hierarchical$^2$ to enforce both constraint families.

The hierarchical decoders introduce no additional trainable variables or loss penalties, but they change the meanings of the native outputs from direct forecasts to residuals. They therefore require loading the corresponding model weights and retraining the model. Their correction errors cannot be determined from the two forecast examples alone because their outputs depend on the residuals learned during retraining. Errors may also accumulate along the recursive chain: an upper quantile depends on the preceding lower quantile, so an error in that lower quantile is propagated to all subsequent upper quantiles. The same propagation can occur from the median toward the lower quantiles.

\noindent\textbf{Joint Projection.}
To evaluate the stage-wise design of KQSP, we introduce Joint Projection as a comparison variant rather than an existing baseline. Like KQSP, JP corrects both crossing types, but it enforces both constraints simultaneously:
\begin{equation}
\widehat{\mathbf Z}^{\mathrm{JP}}
=
\arg\min_{\mathbf V\in\mathcal C_{\mathrm J}}
\sum_{q=1}^{Q}
\sum_{f\in\mathcal F}
(v_{qf}-z_{qf})^2,
\label{eq:joint-projection}
\end{equation}
where $C_{\mathrm J}$ denotes the set satisfying both the quantile and K-line constraints.
Because this is a strictly convex quadratic program with linear constraints, it has a unique global solution. Since KQSP is also feasible,
\begin{equation}
E_{\mathrm{corr}}
\!\left(\widehat{\mathbf Z}^{\mathrm{JP}},\mathbf Z\right)
\leq
E_{\mathrm{corr}}
\!\left(\widehat{\mathbf Z}^{\mathrm{KQSP}},\mathbf Z\right).
\label{eq:joint-error-bound}
\end{equation}
In Table~\ref{tab:correction-examples}, KQSP happens to attain the same correction error as Joint Projection.
Joint Projection guarantees the minimum correction error but solves one coupled problem over all forecasts, even when only one crossing type is present. KQSP instead uses sequential projections, and an already-satisfied stage can be skipped. We include Joint Projection to test whether its theoretical reduction in correction error yields an empirical improvement over KQSP.

\begin{table*}[t]
\caption{Before--after test metrics for the validation-best ANN of each stock. }
\centering
\scriptsize
\setlength{\tabcolsep}{1.8pt}
\label{tab:table-i}
\resizebox{\textwidth}{!}{%
\begin{tabular}{lrrcrrcrrcrrcrrcrrc}
\toprule
& \multicolumn{3}{c}{QCR (\%)}
& \multicolumn{3}{c}{KCR (\%)}
& \multicolumn{3}{c}{AQL}
& \multicolumn{3}{c}{QCE (\%)}
& \multicolumn{3}{c}{MAE}
& \multicolumn{3}{c}{RMSE} \\
Stock
& Before & After & $P$
& Before & After & $P$
& Before & After & $P$
& Before & After & $P$
& Before & After & $P$
& Before & After & $P$ \\
\midrule
BMW
& 82.41 & 0.00 & \goodstars{***}
& 10.76 & 0.00 & \goodstars{***}
& 0.39 & 0.39 & \goodstars{***}
& 3.49 & 3.43 &
& 0.96 & 0.95 & \goodstars{***}
& 1.39 & 1.38 & \goodstars{**} \\

Mercedes
& 86.61 & 0.00 & \goodstars{***}
& 35.70 & 0.00 & \goodstars{***}
& 0.24 & 0.24 & \goodstars{***}
& 3.98 & 3.32 & \goodstars{**}
& 0.60 & 0.59 & \goodstars{***}
& 0.86 & 0.85 & \goodstars{***} \\

Volkswagen
& 100.00 & 0.00 & \goodstars{***}
& 94.75 & 0.00 & \goodstars{***}
& 0.50 & 0.48 & \goodstars{***}
& 10.87 & 11.15 &
& 1.12 & 1.13 & 
& 1.58 & 1.59 & \\
\midrule
BP
& 73.95 & 0.00 & \goodstars{***}
& 16.32 & 0.00 & \goodstars{***}
& 2.22 & 2.21 & \goodstars{***}
& 2.83 & 2.79 &
& 5.43 & 5.42 & \goodstars{*}
& 8.08 & 8.07 & \\

PetroChina
& 44.66 & 0.00 & \goodstars{***}
& 0.55 & 0.00 &
& 0.04 & 0.04 & \goodstars{***}
& 6.70 & 6.70 &
& 0.10 & 0.11 &
& 0.18 & 0.18 & \\

Shell
& 84.21 & 0.00 & \goodstars{***}
& 27.89 & 0.00 & \goodstars{***}
& 10.93 & 10.88 & \goodstars{***}
& 3.65 & 3.72 &
& 26.33 & 26.28 &
& 39.53 & 39.42 & \\
\midrule
Goldman
& 100.00 & 0.00 & \goodstars{***}
& 92.86 & 0.00 & \goodstars{***}
& 3.83 & 3.79 & \goodstars{***}
& 5.83 & 5.59 &
& 9.48 & 9.47 &
& 13.62 & 13.62 & \\

JPMorgan
& 100.00 & 0.00 & \goodstars{***}
& 97.35 & 0.00 & \goodstars{***}
& 1.21 & 1.18 & \goodstars{***}
& 5.98 & 5.30 &
& 3.12 & 2.90 & \goodstars{***}
& 4.17 & 4.01 & \goodstars{***} \\

LSEG
& 100.00 & 0.00 & \goodstars{***}
& 41.84 & 0.00 & \goodstars{***}
& 52.34 & 51.55 & \goodstars{***}
& 6.37 & 6.62 &
& 130.82 & 125.06 & \goodstars{***}
& 217.15 & 211.68 & \goodstars{***} \\
\midrule
Google
& 100.00 & 0.00 & \goodstars{***}
& 78.84 & 0.00 & \goodstars{***}
& 1.45 & 1.42 & \goodstars{***}
& 9.05 & 8.92 &
& 3.71 & 3.47 & \goodstars{***}
& 5.28 & 4.97 & \goodstars{***} \\

Meta
& 100.00 & 0.00 & \goodstars{***}
& 100.00 & 0.00 & \goodstars{***}
& 4.88 & 4.64 & \goodstars{***}
& 17.27 & 16.40 &
& 11.95 & 11.44 & \goodstars{***}
& 16.54 & 16.06 & \goodstars{***} \\

Microsoft
& 80.42 & 0.00 & \goodstars{***}
& 20.90 & 0.00 & \goodstars{***}
& 1.94 & 1.93 & \goodstars{***}
& 2.68 & 2.75 &
& 4.73 & 4.71 &
& 7.09 & 7.08 & \\
\midrule
Improved or n.s.
& & & 12/12
& & & 12/12
& & & 12/12
& & & 12/12
& & & 12/12
& & & 12/12 \\
\bottomrule
\end{tabular}%
}
\end{table*}

\begin{table*}[t]
\centering
\footnotesize
\caption{Comparison of crossing baselines using the ANNs. Hierarchical$^{1}$ enforces quantile order only, Hierarchical$^{2}$ enforces OHLC validity only, and Hierarchical$^{3}$ enforces both constraints.}
\label{tab:table-ii}
\resizebox{\textwidth}{!}{%
\begin{tabular}{lrrrrrrrrrr}
\toprule
Method
& QCR (\%)
& KCR (\%)
& AQL
& $P$
& Max AQL
& $P$
& Mean corr.
& $P$
& Max corr.
& $P$ \\
\midrule
No Constraint
& 87.69
& 51.48
& 6.66
& \badstars{***}
& 67.14
& \badstars{**}
& --
& --
& --
& -- \\

Cumulative Max.
& 0.00
& 0.00
& 6.57
&
& 66.36
&
& 1.07
& \badstars{***}
& 43.29
& \badstars{***} \\

Reordering
& 0.00
& 0.00
& 6.56
& 
& 66.97
& 
& 1.60
& \badstars{***}
& 33.59
& \badstars{***} \\

Hierarchical$^{1}$
& 0.00
& 21.95
& 6.96
& \badstars{*}
& 57.96
&
& 9.26
& \badstars{***}
& 81.78
& \badstars{***} \\

Hierarchical$^{2}$
& 94.85
& 0.00
& 7.26
& \badstars{**}
& 64.80
&
& 11.55
& \badstars{***}
& 99.98
& \badstars{***} \\

Hierarchical$^{3}$
& 0.00
& 0.00
& 8.53
& \badstars{***}
& 64.59
&
& 14.90
& \badstars{***}
& 125.41
& \badstars{***} \\

Joint Projection
& 0.00
& 0.00
& 6.56
&
& 66.06
&
& 0.91
&
& 22.94
& \\

Proposed KQSP
& 0.00
& 0.00
& 6.56
&
& 66.06
&
& 0.91
&
& 22.94
& \\
\bottomrule
\end{tabular}%
}
\end{table*}

\subsection{Forecasting Models}
\label{sec:models}


\noindent\textbf{ANN.}
The artificial neural network (ANN) is trained from scratch for each stock and serves as the reference model. It represents a conventional task-specific approach against which the zero-shot foundation models are compared later. For each stock, we sample 100 ANN configurations.
The number of hidden layers is sampled from $\{1,2,3,4,5\}$; units per layer from the integers $[2,512]$; learning rate log-uniformly from $[10^{-5},10^{-2}]$; dropout uniformly from $[0,0.9]$.
Each model uses the nine quantiles $\mathcal T=\{0.1,0.2,\ldots,0.9\}$, batch size of 256, and 500 epochs.
The validation-best configuration is selected separately for each stock, and its untouched test split is used for Table~\ref{tab:table-i}.

\noindent\textbf{TimesFM.}
Developed by Google Research, TimesFM is pretrained on large-scale real-world and synthetic time series from diverse domains~\cite{das2024timesfm}. We evaluate the original TimesFM and the newer 200-million-parameter TimesFM 2.5.

\noindent\textbf{Chronos.}
Developed by Amazon, the Chronos family provides generic probabilistic forecasting based on public and synthetic pretraining data~\cite{ansari2024chronos,ansari2025chronos2}. We evaluate both Chronos and Chronos-2, which support multivariate and covariate-informed forecasting.

\noindent\textbf{Moirai.}
Moirai is a universal time-series foundation model pretrained on the Large-scale Open Time Series Archive (LOTSA)~\cite{woo2024moirai}. LOTSA contains more than 27 billion observations collected from nine application domains.

\noindent\textbf{Time-MoE.}
Time-MoE is a sparse mixture-of-experts foundation model designed to investigate scaling and computational efficiency in time-series forecasting~\cite{shi2025timemoe}. It is pretrained on Time-300B, which contains more than 300 billion time points from nine domains.

\noindent\textbf{TabPFN.}
TabPFN-TS adapts a tabular foundation model to zero-shot time-series forecasting~\cite{hollmann2022tabpfn}. Because the underlying model is pretrained entirely on synthetic data, its pretraining corpus does not contain observations from real-world forecasting benchmarks.

\noindent\textbf{Kronos.}
Kronos is a finance-specific foundation model pretrained on more than 12 billion OHLCV candlestick records from 45 global exchanges and seven temporal granularities~\cite{shi2026kronos}. 
The Kronos pretraining corpus ends in June 2024, whereas the shared test period used by Kronos and the fully trained ANN spans January 17, 2025 to July 23, 2026. Therefore, no test observations overlap the Kronos pretraining period.

\noindent\textbf{Input and Output Specification.}
The univariate models TimesFM, Chronos, and Time-MoE receive the historical sequence of each OHLC target separately. TimesFM 2.5 and TabPFN receive each target history together with the remaining OHLCV channels as lag covariates, whereas Chronos-2, Moirai, and Kronos receive multivariate historical sequences. TimesFM, TimesFM 2.5, Chronos-2, and TabPFN directly provide probabilistic outputs. Chronos, Moirai, and Kronos construct quantiles from sampled future paths, while Time-MoE converts its point forecast into quantiles using past one-step residuals.


\section{Evaluation Metrics}
\label{sec:metrics}

Let $i\in\{1,\ldots,N\}$ index the $N$ test samples, let $y_{if}$ be the realized value of feature $f$, and let $z_{iqf}$ be its raw quantile forecast.

\noindent\textbf{Quantile Crossing Rate (QCR)} is the percentage of samples with at least one quantile crossing,
\begin{equation}
  \mathrm{QCR}=\frac{100}{N}\sum_{i=1}^{N}\mathbf1\!\left\{\exists(q,f):z_{iqf}>z_{i,q+1,f}\right\}.
  \label{eq:qcr}
\end{equation}
\noindent\textbf{K-Line Crossing Rate (KCR)} is the percentage of samples with at least one K-line crossing,
\begin{equation}
  \begin{split}
  \mathrm{KCR}=\frac{100}{N}\sum_{i=1}^{N}\mathbf1\!\{\exists q:\;&z_{iqH}<\max(z_{iqO},z_{iqC})\ \text{or}\\
  &z_{iqL}>\min(z_{iqO},z_{iqC})\}.
  \end{split}
  \label{eq:KCR}
\end{equation}
Here $\mathbf1\{\cdot\}$ is the indicator function.
The pinball loss at quantile level $\tau$ for error $e$ is
\begin{equation}
  \rho_{\tau}(e)=e\left(\tau-\mathbf1\{e<0\}\right).
  \label{eq:pinball}
\end{equation}
\noindent\textbf{Average Quantile Loss (AQL)} averages quantile loss over samples, quantiles, and OHLC features,
\begin{equation}
  \mathrm{AQL}=\frac{1}{4QN}\sum_{i=1}^{N}\sum_{q=1}^{Q}\sum_{f\in\mathcal F}\rho_{\tau_q}(y_{if}-z_{iqf}).
  \label{eq:aql}
\end{equation}
Max AQL is the largest sample-level AQL.

\noindent\textbf{Quantile coverage error (QCE)}
 averages the absolute empirical-coverage deviation,
\begin{equation}
  \mathrm{QCE}=\frac{100}{Q}\sum_{q=1}^{Q}\left|\frac{1}{4N}\sum_{i=1}^{N}\sum_{f\in\mathcal F}\mathbf1\{y_{if}\leq z_{iqf}\}-\tau_q\right|.
  \label{eq:qce}
\end{equation}
We use standard RMSE and MAE for pointwise evaluation.

\section{Case Study}
\label{sec:case-study}
\subsection{Effectiveness in the Full-Training Setting}

We first use the fully trained ANN to determine whether KQSP resolves both crossing issues without compromising predictive accuracy.
The model chooses the optimal historical input length from $\{1,3,7,30\}$ trading days based on validation data. 
Within each stock, Table~\ref{tab:table-i} compares paired test-day contributions \underline{before} and \underline{after} KQSP using a two-sided sign-flip test.
Green stars denote significant improvement, red stars denote significant deterioration, and a blank denotes no significant difference. One, two, and three stars indicate $p<0.05$, $p<0.01$, and $p<0.001$, respectively. The last row counts improvement or no significant difference across stocks.

First, the raw ANN forecasts exhibit severe crossings: the quantile crossing rate (QCR) ranges from 44.66\% to 100.00\%, while the K-line crossing rate (KCR) ranges from 0.55\% to 100.00\%.
Second, KQSP reduces both crossing rates to zero for every stock.
Third, although KQSP is designed to enforce consistency rather than improve predictive accuracy, it significantly improves AQL for all 12 stocks ($p<0.001$) and improves several MAE and RMSE results.
A plausible explanation is that the violations partly reflect estimation noise: by projecting the forecasts onto economically valid OHLC and quantile constraints, KQSP removes infeasible variation while making the smallest possible correction.
This shape-constrained correction can therefore move inconsistent forecasts closer to the realized values without changing already valid components unnecessarily.

\begin{table*}[t]
\centering
\scriptsize
\setlength{\tabcolsep}{1.8pt}
\caption{Before--after results aggregated across stocks. }
\label{tab:table-iii}
\resizebox{\textwidth}{!}{%
\begin{tabular}{lrrcrrcrrcrrcrrcrrc}
\toprule
& \multicolumn{3}{c}{QCR (\%)}
& \multicolumn{3}{c}{KCR (\%)}
& \multicolumn{3}{c}{AQL}
& \multicolumn{3}{c}{QCE (\%)}
& \multicolumn{3}{c}{MAE}
& \multicolumn{3}{c}{RMSE} \\
Model
& Before & After & $P$
& Before & After & $P$
& Before & After & $P$
& Before & After & $P$
& Before & After & $P$
& Before & After & $P$ \\
\midrule

ANN
& 87.69 & 0.00 & \goodstars{***}
& 51.48 & 0.00 & \goodstars{***}
& 6.66 & 6.56 & \goodstars{***}
& 6.56 & 6.39 &
& 16.53 & 15.96 & \goodstars{**}
& 26.29 & 25.74 & \goodstars{**} \\

TimesFM
& 1.07 & 0.00 & \goodstars{**}
& 52.66 & 0.00 & \goodstars{***}
& 8.37 & 8.33 & \goodstars{***}
& 4.16 & 4.12 &
& 20.84 & 20.73 & \goodstars{***}
& 31.19 & 31.05 & \goodstars{***} \\

TimesFM 2.5
& 3.14 & 0.00 & \goodstars{***}
& 62.46 & 0.00 & \goodstars{***}
& 8.01 & 7.83 & \goodstars{***}
& 2.33 & 2.27 & \goodstars{*}
& 19.88 & 19.36 & \goodstars{***}
& 29.89 & 29.08 & \goodstars{***} \\

Chronos
& 0.00 & 0.00 &
& 83.09 & 0.00 & \goodstars{***}
& 7.79 & 7.68 & \goodstars{***}
& 2.28 & 2.18 & \goodstars{*}
& 19.16 & 18.95 & \goodstars{***}
& 29.51 & 29.24 & \goodstars{***} \\

Chronos-2
& 0.04 & 0.00 &
& 46.27 & 0.00 & \goodstars{***}
& 7.71 & 7.68 & \goodstars{***}
& 1.90 & 1.93 &
& 19.17 & 19.09 & \goodstars{***}
& 28.98 & 28.90 & \goodstars{***} \\

Moirai
& 0.00 & 0.00 &
& 69.00 & 0.00 & \goodstars{***}
& 8.15 & 8.12 & \goodstars{***}
& 3.49 & 3.46 &
& 20.05 & 20.01 & \goodstars{***}
& 30.10 & 30.05 & \goodstars{***} \\

Time-MoE
& 0.00 & 0.00 &
& 76.12 & 0.00 & \goodstars{***}
& 8.02 & 7.92 & \goodstars{***}
& 3.86 & 3.70 & \goodstars{**}
& 19.76 & 19.54 & \goodstars{***}
& 29.70 & 29.44 & \goodstars{***} \\

TabPFN
& 0.00 & 0.00 &
& 63.18 & 0.00 & \goodstars{***}
& 6.90 & 6.86 & \goodstars{***}
& 1.37 & 1.36 &
& 16.87 & 16.78 & \goodstars{***}
& 26.56 & 26.41 & \goodstars{***} \\

Kronos
& 0.00 & 0.00 &
& 2.52 & 0.00 & \goodstars{***}
& 6.55 & 6.55 & \goodstars{***}
& 7.64 & 7.63 &
& 15.73 & 15.73 & \goodstars{*}
& 24.92 & 24.91 & \goodstars{*} \\

\bottomrule
\end{tabular}%
}
\end{table*}

\begin{table*}[t]
\centering
\footnotesize
\caption{Upper-triangular pairwise comparison of post-KQSP AQL across stocks. Diagonal entries are mean AQL. Each upper cell is a two-sided paired Wilcoxon test between its row and column models: green stars favor the row model, red stars favor the column model, and a blank denotes no significant difference.}
\label{tab:table-iv}
\resizebox{0.95\textwidth}{!}{%
\begin{tabular}{lccccccccc}
\toprule
Model & ANN & TimesFM & TimesFM 2.5 & Chronos & Chronos-2 & Moirai & Time-MoE & TabPFN & Kronos \\
\midrule
ANN
& \textbf{6.56}
& \goodstars{***}
& \goodstars{***}
& \goodstars{**}
& \goodstars{***}
& \goodstars{***}
& \goodstars{***}
& \goodstars{**}
& \\

TimesFM
&
& \textbf{8.33}
& \badstars{***}
& \badstars{***}
& \badstars{***}
& \badstars{***}
& \badstars{***}
& \badstars{***}
& \badstars{***} \\

TimesFM 2.5
&
&
& \textbf{7.83}
& \badstars{**}
& \badstars{***}
& \goodstars{*}
& \goodstars{***}
& \badstars{***}
& \badstars{***} \\

Chronos
&
&
&
& \textbf{7.68}
&
& \goodstars{***}
& \goodstars{***}
& \badstars{***}
& \badstars{***} \\

Chronos-2
&
&
&
&
& \textbf{7.68}
& \goodstars{***}
& \goodstars{***}
& \badstars{***}
& \badstars{***} \\

Moirai
&
&
&
&
&
& \textbf{8.12}
&
& \badstars{***}
& \badstars{***} \\

Time-MoE
&
&
&
&
&
&
& \textbf{7.92}
& \badstars{***}
& \badstars{***} \\

TabPFN
&
&
&
&
&
&
&
& \textbf{6.86}
& \badstars{***} \\

Kronos
&
&
&
&
&
&
&
&
& \textbf{6.55} \\
\bottomrule
\end{tabular}%
}
\end{table*}

\subsection{Comparison of Existing Crossing Solutions}

No Constraint, Cumulative Maximum, Reordering, Joint Projection, and KQSP are applied directly to the same saved forecasts without further training.
The three hierarchical variants load the same base ANN weights and are retrained using the same hyperparameters.
Table~\ref{tab:table-ii} compares each baseline with \underline{KQSP} using a two-sided paired Wilcoxon test across the stocks.
Mean and maximum correction measure absolute changes in native price units and are particularly important when the original forecasts are trusted and only minimal corrections are desired.

First, Cumulative Maximum, Reordering,  Hierarchical$^3$, Joint Projection, and KQSP eliminate both crossing types. Especially, KQSP achieves a significantly lower AQL than No Constraint ($p<0.001$), showing that consistency can be imposed while even \textbf{improving} the predictive accuracy with statistical significance ($p<0.001$).
Second, Cumulative Maximum and Reordering have no significant AQL difference from KQSP. Nevertheless, both produce significantly larger mean and maximum correction errors ($p<0.001$).
This distinction is critical when the forecasting model is trusted, and the objective is to resolve crossings with the smallest possible modification.
Third, the hierarchical AQL values follow Hierarchical$^3$ $>$ Hierarchical$^2$ $>$ Hierarchical$^1$, with Hierarchical$^3$ performing worst.
The deterioration caused by combining both recursive constructions is consistent with error propagation across the quantile and OHLC hierarchies.
Lastly, Joint Projection does not differ significantly from the proposed KQSP across any metric (all $p>0.05$), as indicated by the absence of significance stars in the table. This shows that KQSP achieves empirically equivalent correction performance while offering greater stage-wise flexibility.

\subsection{Sensitivity to Different Models}

To mitigate model-specific bias and investigate whether KQSP can consistently improve AQL across different forecasting models, we compare the fully trained ANN with zero-shot foundation models using identical test observations and quantile levels.
Each model's raw forecast is evaluated before and after the KQSP operator.
Table~\ref{tab:table-iii} reports paired before--after results aggregated across stocks. Before is the raw forecast and After is the same forecast after KQSP. Each $P$ cell is a two-sided paired Wilcoxon test of all stocks Before and After metrics.  
Table~\ref{tab:table-iv} compares post-KQSP AQL between models.


First, KQSP removes all quantile and K-line crossings for every model. Surprisingly, the results confirm that KQSP can consistently improve predictive metrics, including AQL, MAE, and RMSE, with statistical significance.
Second, Table~\ref{tab:table-iv} shows no significant AQL difference between the fully trained ANN and Kronos, indicating strong zero-shot probabilistic performance from the finance-specific Kronos model. By contrast, the generic foundation models do not match the simple ANN on this dataset.
One possible explanation is that these models are pretrained primarily on broad and generic time-series data and may not sufficiently capture stock-specific price dynamics.
However, this experiment is \textbf{not} intended to establish a universal ranking of forecasting models.
Its purpose is to determine whether KQSP eliminates both crossing issues without degrading predictive accuracy, regardless of whether the forecasts are produced by a strong or weak model.

\section{Conclusion}
\label{sec:conclusion}

This study introduced KQSP, a parameter-free method for reconciling probabilistic K-line forecasts without retraining the forecasting model.  Each stage of KQSP minimizes the squared correction, allowing KQSP to remove both crossing types while preserving the original forecasts as closely as possible.
Three case studies establish the effectiveness of KQSP. First, for fully trained ANNs across diverse stocks, KQSP reduces both QCR and KCR to zero and significantly improves AQL for every stock. Second, KQSP achieves predictive accuracy comparable to other strong baselines while producing significantly smaller mean and maximum corrections. Third, experiments involving zero-shot foundation models confirm that KQSP eliminates crossings and can even improve partial forecasting metrics across different models. 
A limitation of KQSP is that it applies the minimum correction required for consistency, regardless of whether the original forecasts are strong or weak. Future work should investigate methods that not only eliminate crossings but also improve predictive accuracy when the original forecasts are inaccurate.


\bibliographystyle{ACM-Reference-Format}
\bibliography{references}

\end{document}